\documentclass[acmtog,nonacm]{acmart}

\makeatletter
\let\@authorsaddresses\@empty
\makeatother

\usepackage{booktabs}
\usepackage{enumitem}

\usepackage{xcolor}

\usepackage{multirow}
\usepackage{bigstrut}
\usepackage{subcaption}
\usepackage{tikz}

\usepackage{multicol}
\usepackage[ruled]{algorithm2e}
\usepackage{wrapfig}
\usepackage{graphicx}
\usepackage{lipsum}
\usepackage{enumitem}
\usepackage{microtype}

\SetAlFnt{\small}
\SetAlCapFnt{\small}
\SetAlCapNameFnt{\small}
\SetAlCapHSkip{0pt}

\newcommand{\zmin}{z_{\min}}
\newcommand{\zmax}{z_{\max}}

\acmJournal{TOG}

\begin{document}

\title{Oblique-MERF: Revisiting and Improving MERF for Oblique Photography}

    \author{Xiaoyi Zeng}
    \affiliation{%
      \institution{University of Science and Technology of China}
      \country{China}
    }
    
    \author{Kaiwen Song}
    \affiliation{%
      \institution{University of Science and Technology of China}
      \country{China}
    }
    
    \author{Leyuan Yang}
    \affiliation{%
      \institution{University of Science and Technology of China}
      \country{China}
    }
    
    \author{Bailin Deng}
    \affiliation{%
      \institution{Cardiff University}
      \country{United Kingdom}
    }
    
    \author{Juyong Zhang}
    \authornote{Corresponding author (\href{mailto:juyong@ustc.edu.cn}{juyong@ustc.edu.cn}).}
    \affiliation{%
      \institution{University of Science and Technology of China}
      \country{China}
    }

\begin{teaserfigure}
  \includegraphics[width=\textwidth]{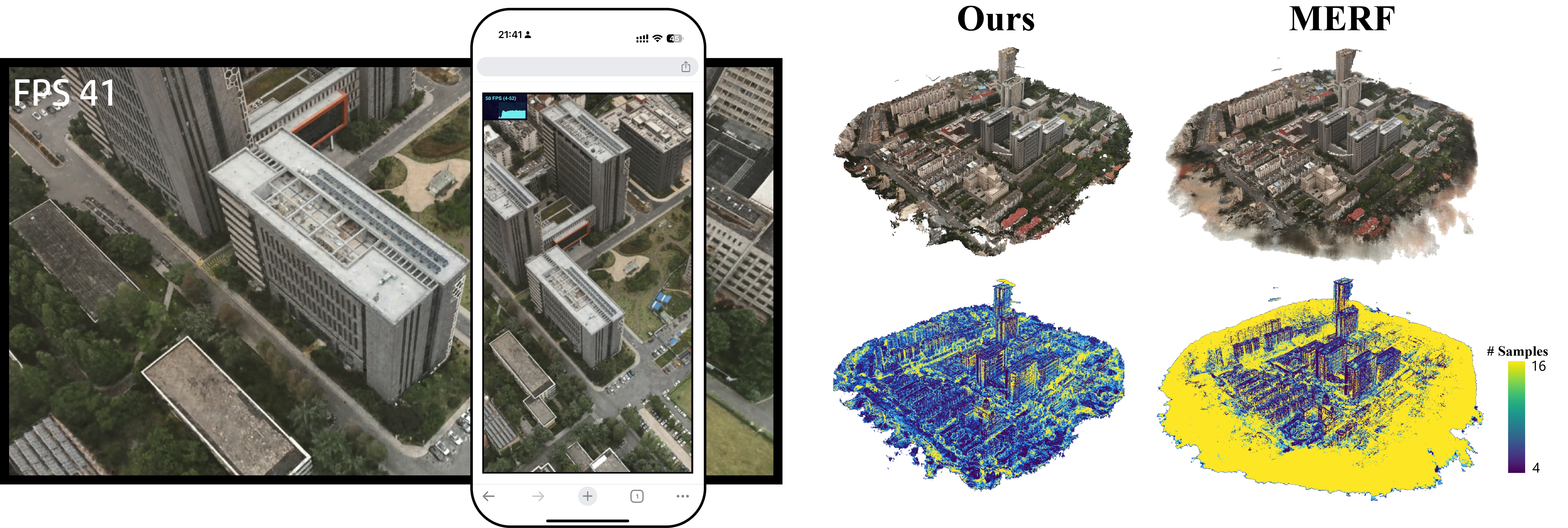}
  \caption{Oblique-MERF enables real-time synthesis of novel views for oblique photography on diverse commodity devices, including tests on an NVIDIA GTX 1650 and an iPhone 14 Pro Max (left). We introduce a novel compact sampling representation that markedly reduces sampling points and emphasizes significant regions, enhancing rendering quality and increasing frame rates. We present the rendering output and the visualization of the number of samples (right).}
  \label{fig:teaser}
\end{teaserfigure}

\begin{abstract}

Neural implicit fields have established a new paradigm for scene representation, with subsequent work achieving high-quality real-time rendering. However, reconstructing 3D scenes from oblique aerial photography presents unique challenges, such as varying spatial scale distributions and a constrained range of tilt angles, often resulting in high memory consumption and reduced rendering quality at extrapolated viewpoints. In this paper, we enhance MERF to accommodate these data characteristics by introducing an innovative adaptive occupancy plane optimized during the volume rendering process and a smoothness regularization term for view-dependent color to address these issues. Our approach, termed Oblique-MERF, surpasses state-of-the-art real-time methods by approximately 0.7 dB, reduces VRAM usage by about $40\%$, and achieves higher rendering frame rates with more realistic rendering outcomes across most viewpoints.
\end{abstract}



\maketitle

\section{Introduction}

Reconstructing a 3D scene for high-fidelity rendering from freely chosen viewpoints has been a longstanding challenge in computer graphics. Neural Radiance Field (NeRF)~\cite{mildenhall2020nerf} accomplishes this via a novel implicit representation parameterized by multi-layer perceptrons (MLP). In the realm of large-scale scene reconstruction for oblique aerial photography datasets, various works have sought to improve upon NeRF by employing strategies such as spatial partitioning~\cite{reiser2021kilonerf,Tancik_2022_CVPR_Block-NeRF,mi2023switchnerf}, advanced sampling methods~\cite{adaptiveshells2023,turki2023hybridnerf}, and efficient data structures~\cite{mueller2022instant,SunSC22DVGO,yu2021plenoctrees,yu_and_fridovichkeil2021plenoxels}. Furthermore, recent works bake models into meshes with optimized textures~\cite{chen2022mobilenerf,yariv2023bakedsdf,tang2022nerf2mesh} or feature grids and planes~\cite{hedman2021snerg,Reiser2023SIGGRAPHMERF}, to enable interactive rendering frame rates on commercial devices.

However, these advancements have not been specifically tailored to the unique characteristics of oblique aerial photography data. When applied to such large-scale scenes, they incur high memory footprints and fail to deliver high-fidelity rendering from all perspectives. Specifically, oblique aerial photography datasets often cover vast areas, spanning hundreds of thousands of square meters yet only extending to a few hundred meters of vertical height. The commonly used cubic grid~\cite{mueller2022instant} has a high complexity of storage and struggles to adapt to such data, leading to inefficient use of storage and computational resources. Additionally, such datasets might lead to artifacts such as floaters due to inadequate constraints on the sampling space. Moreover, instead of a complete 360-degree view, these scenes are often captured from a constrained range of pitch angles relative to the ground and may exhibit abnormal highlights and shadows in viewpoints not covered during training. In this work, we enhance the performance and quality of current NeRF-based real-time rendering methods on oblique aerial photography datasets, focusing on two key aspects: the representation of the sampling space and the issue of color extrapolation.

Current prevalent methods, such as 3D grids or implicit proposal sampling networks~\cite{mueller2022instant,barron2022mipnerf360}, may face significant limitations when dealing with such vast scenes as shown in Tab.~\ref{tab:sampling representations}. On the one hand, the cubic storage complexity  of grid-based representations limits their ability to represent sampling spaces at high resolutions; moreover, they often rely on optimization methods that are not inherently tied to rendering quality. These limitations can hinder the precise conveyance of occupancy information and lead to redundant sample points that slow down the rendering pipeline and affect the rendering quality. On the other hand, network-based representations are computationally expensive, as they require evaluating numerous candidate points to determine the final sample points. To overcome these drawbacks, we propose a novel sampling strategy that models the occupancy space as a sandwiched region between two height field surfaces conforming to the ground.
Furthermore, we integrate this adaptive representation with volume rendering to ensure awareness of the photometric loss. Thanks to our explicit representation of the sampling space, features of the occupied region can be directly extracted for real-time rendering. This eliminates the need for tens of hours of baking.

\begin{table}[t]
\caption{Difference among sampling representations.}
  \centering
  \setlength{\tabcolsep}{3pt} 
  \renewcommand{\arraystretch}{0.7} 
    \begin{tabular}{cccc}
    \toprule
    \multirow{2}[2]{*}{} & Memory & Query & Loss \bigstrut[t]\\
          & efficient & speed & aware \bigstrut[b]\\
    \midrule
    Occupancy Grid~\cite{mueller2022instant} & $\times$     & $\checkmark$     & $\times$ \bigstrut[t]\\
    Proposal MLP~\cite{barron2022mipnerf360} & $\checkmark$     & $\times$    & $\checkmark$ \\
    Occupancy Plane (Ours) & $\checkmark$     & $\checkmark$     & $\checkmark$ \bigstrut[b]\\
    \bottomrule
    \end{tabular}%
  
  \label{tab:sampling representations}%
\end{table}

In addition, we address the extrapolation issue arising from the limited range of view directions in the captured training set. In the setting of extrapolation novel view, the rendering results often exhibit aberrant colors due to the lack of supervision. As such anomalies primarily result from the non-smooth behavior of specular color with respect to varying viewpoints, we introduce a novel smoothing term suitable for oblique photography to regularize color dependence on observation directions. Compared to directly constraining the Lipschitz continuity in neural networks~\cite{lipmlp}, our approach focuses on the change of the specular component with the viewing directions and exhibits stronger generalizability and robustness, offering a more natural handling of color variations related to viewpoints during rendering from most perspectives.

In summary, our primary contributions are:
\begin{itemize}[leftmargin=*]
    \item We propose a novel sampling strategy that seamlessly integrates training and baking processes by optimizing an explicit occupancy representation. The adaptive structure is aware of the photometric loss and the regularization term to convey occupancy information at the least memory cost accurately.
    \item Utilizing the smooth nature of specular reflections, we present a new regularization approach that significantly enhances the rendering quality of novel extrapolated viewpoints, providing smoother and more realistic visual outcomes.
    \item Extensive experiments confirm that our method enhances the PSNR by 0.7 dB, decreases VRAM usage by $40\%$, and boosts the frame rate by $35\%$ and $300\%$ for low-altitude and high-altitude viewpoints, respectively, compared to the baseline. Additionally, our technique for smoothing view-dependent color significantly improves PSNR by 0.52 dB in extrapolation scenarios.
\end{itemize}

\section{Related Work}

Our work primarily focuses on large-scale scene reconstruction and real-time rendering. Below, we review domains related to our work, including large-scale scene representation, real-time rendering, sampling strategy, and regularization of the radiance field.

\paragraph{3D Reconstruction for Oblique Photography.} To reconstruct large-scale scenes, a common approach involves using drones equipped with one or more cameras to capture high-altitude flight data across the scene, known as oblique photography. Traditional 3D modeling techniques include the use of Structure-from-Motion(SfM) pipelines~\cite{visualsfm,sfm,sfm2,colmap} to estimate camera poses and obtain sparse point clouds, followed by surface reconstruction through dense multi-view stereo~\cite{mvs1,mvs2,mvs3}. These methods rely on manual operations to acquire fine textures and geometry, and struggle to reconstruct view-dependent colors. With the advent of Neural Radiance Fields, neural rendering for novel view synthesis has been increasingly applied to large-scale scene reconstruction. The original NeRF~\cite{mildenhall2020nerf} represents the scene as an MLP, which maps positional encodings of spatial locations and directions to attributes such as color and volumetric density, and utilizes volume rendering principles for realistic rendering outcomes. Recently, there has been a notable increase in research efforts to adapt NeRF for large-scale scenes such as oblique Photography datasets. Some approaches~\cite{Turki_2022_CVPR_Mega-NERF,Tancik_2022_CVPR_Block-NeRF,mi2023switchnerf,song2023cityonweb} adopt a divide-and-conquer strategy, segmenting the scene into chunks to perform parallel reconstruction, and then merging them to represent the entire scene holistically. GridNeRF~\cite{xu2023gridguided} combines a multi-resolution ground feature plane representation with vanilla NeRF incorporating position-encoded inputs, enabling a collaborative learning process for rendering. Moreover, BungeNeRF~\cite{xiangli2022bungeenerf} employs residual networks to learn multi-scale features, fitting scenes with dramatic changes in altitude.

\paragraph{Real-Time Rendering.} Many methods accelerate rendering by reducing the volume of network queries~\cite{neff2021donerf,kurz-adanerf2022,song2019autoint} or by decomposing larger MLPs to facilitate parallel processing~\cite{reiser2021kilonerf}. Some works~\cite{mueller2022instant,SunSC22DVGO,yu2021plenoctrees,yu_and_fridovichkeil2021plenoxels,TensoRF} reach it by introducing explicit structures, such as regular 3D voxel grids or octrees, storing features to replace partial or complete network. Other approaches seek to circumvent extensive network queries by baking models into explicit structures. For example, SNeRG~\cite{hedman2021snerg} extracts a sparse 3D voxel grid that stores density, diffuse color, and specular color. During rendering, each ray only needs to traverse a small MLP once. On the contrary, certain works~\cite{chen2022mobilenerf,yariv2023bakedsdf,tang2022nerf2mesh,vmesh,Nres,Duplex,neuralAssets} utilize optimized surfaces with textures to represent scenes, integrating into modern computer graphics rendering pipelines. While achieving interactive frame rates on commercial devices, they fall short in rendering quality and memory efficiency for large-scale scenes. MERF~\cite{Reiser2023SIGGRAPHMERF} employs a memory-efficient triplane combined with a sparse grid to represent the features of spatial points, achieving a significant reduction in memory consumption without compromising quality.

\paragraph{Sampling in Rendering.} Various methods enhance rendering efficiency by refining the sampling strategy. NeRF~\cite{mildenhall2020nerf} and Mip-NeRF 360~\cite{barron2022mipnerf360} employ a coarse-to-fine strategy to concentrate on significant regions. DDNeRF~\cite{Dadon_2023_WACV} adopts the Gaussian function instead of a piecewise-constant probability density function, achieving accurate density representation. DONeRF~\cite{neff2021donerf} and ENeRF~\cite{lin2022efficient-nerf} use depth information to reduce the number of sampling points. NeuSample~\cite{fang2021neusample} directly maps rays to sampling points through a single inference. In contrast to the network-based sampling method, Instant-NGP~\cite{mueller2022instant} skips empty space by explicit multi-resolution occupancy grids. While Adaptive Shells~\cite{adaptiveshells2023} and HybridNeRF~\cite{turki2023hybridnerf} refine the sampling interval size across different spatial locations by optimizing the spatially-varying parameter, compressing the sampling area.  

\paragraph{Regularizations of Neural Fields.} In addition to these advancements, subsequent work on NeRF has introduced various regularization terms to enhance its performance. Indeed, in the realm of neural fields, numerous studies have incorporated regularization terms to encourage the smoothness of networks, such as penalties on the norms of Jacobians and Hessians~\cite{jacob,moosavidezfooli2018robustness} or encouragement of smaller Lipschitz constants for weights~\cite{lipmlp}. In NeRF-related works, Plenoxels~\cite{yu_and_fridovichkeil2021plenoxels} proposes a total variation loss to minimize differences in features among adjacent voxels, and RegNeRF~\cite{Niemeyer2021Regnerf} aims to encourage the continuity of volumetric density changes. Additionally, sparse regularization~\cite{yu2021plenoctrees,Reiser2023SIGGRAPHMERF} is employed in many real-time rendering schemes to eliminate irrelevant features.

\begin{figure*}[h]
  \centering
  \includegraphics[width=\linewidth]{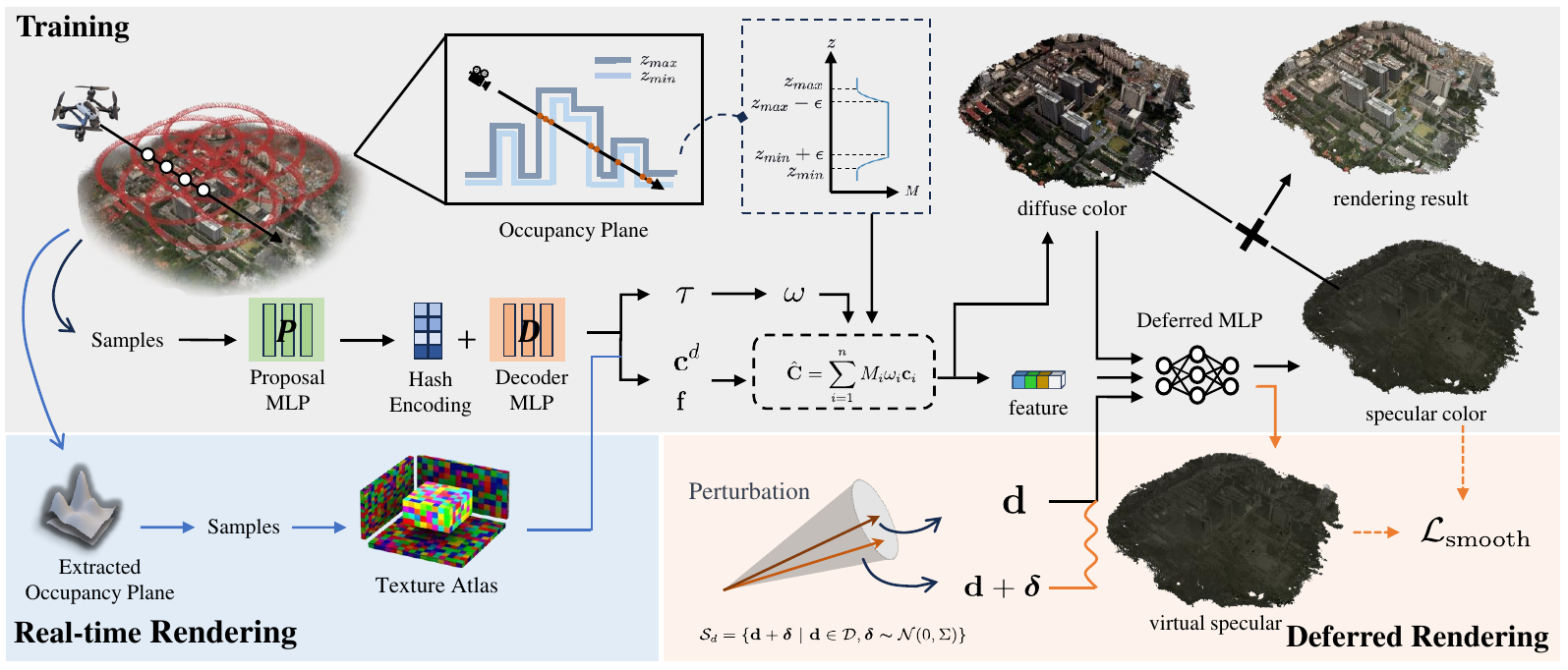}
  \caption{ Overview of our Oblique-MERF pipeline. During training, we introduce a 2D plane to represent the occupied space as a sandwiched region between two height field surfaces. For sampling points on rays, occupancy masks retrieved from the occupancy plane, are used as multipliers in the volume rendering process(section~\ref{section:4.1}). Additionally, we incorporate a smoothness regularization for view-dependent color to minimize variations in specular color with viewing direction(section~\ref{section:4.2}). Post-training, spatial occupancy information is directly extracted from the occupancy plane, and corresponding features are stored for real-time rendering(section~\ref{section:4.4}).}
  \label{fig:pipeline}
\end{figure*}

\section{Background}

NeRF~\cite{mildenhall2020nerf} employs an MLP to represent a scene as a continuous volumetric function $\mathcal{F}:(\mathbf{p},\mathbf{d}) \mapsto (\sigma, \mathbf{c})$, mapping positional encodings of 3D points $\mathbf{p} \in \mathbb{R}^{3}$ and normalized directions $\mathbf{d} \in \mathbb{S}^{2}$ to volumetric density $\tau(\mathbf{p})$ and color $\mathbf{c}(\mathbf{p},\mathbf{d})$. 

To render the corresponding color of a pixel, a ray $\mathbf{r} = \mathbf{o}+t\mathbf{d} $ is first emitted from the origin $\mathbf{o}$ along view directions $\mathbf{d}$, where dozens of points $\{\mathbf{p}_i = \mathbf{o}+t_{i}\mathbf{d}\ |\ i=1,\ldots,n,t_{i}<t_{i+1}\}$ are sampled to estimate their volumetric density and features. These estimates are integrated through the numerical integral form of volumetric rendering to synthesize the final color~\cite{max}:
\begin{equation} \label{alpha}
  \hat{\mathbf{C}}(\mathbf{r}) = \sum\nolimits_{i=1}^{n}\omega_{i}\mathbf{c}_{i},\quad\omega_{i} = T_{i}\alpha_{i},
\end{equation}
where 
\begin{math}
    \alpha_{i} = (1-e^{-\tau_{i}\delta_{i}})
\end{math}
is the opacity of the sample $\mathbf{p}_{i}$, with
\begin{math}
  \delta_{i} = t_{i+1}-t_{i}
\end{math}
being the distance between adjacent samples.
\begin{math}
  T_{i} = \prod_{j=1}^{i-1}(1-\alpha_{j})
\end{math}
is the accumulated transmittance from $\mathbf{p}_{1}$ to $\mathbf{p}_{i-1}$. Subsequently, the MLP is optimized by minimizing the mean squared error between the predicted colors $\{\hat{\mathbf{C}}(\mathbf{r})\}$ along rays emitted from the camera and the groud-truth colors from the input images.

MERF~\cite{Reiser2023SIGGRAPHMERF} employs a low-resolution voxel grid $\mathbf{V}\in \mathbb{R}^{L\times L\times L\times 8}$ and a high-resolution triplane $\{\mathbf{P}_{i} \in \mathbb{R}^{R\times R\times 8}\ |\ i = x,y,z\}$ to represent the scene. For each sample $\mathbf{p}$, the features are obtained by adding the results of trilinear interpolation on the sparse grid and bilinear interpolation on the three planes, respectively:
\begin{equation}
    \mathbf{t}(\mathbf{p}) = \mathbf{V}(\mathbf{p})+\mathbf{P}_{x}(\mathbf{p})+\mathbf{P}_{y}(\mathbf{p}) + \mathbf{P}_{z}(\mathbf{p}).
\end{equation}

Similar to SNeRG~\cite{hedman2021snerg}, to disentangle diffuse color and specular color, the feature is split into three components $\mathbf{t}(\mathbf{p}) = [\tilde{\tau}, \tilde{\mathbf{c}}^{d}, \tilde{\mathbf{f}}]$. And exponential $\exp(\cdot)$ and sigmoid activation function $\sigma(\cdot)$ are applied to obtain separately volumetric density $\tau \in \mathbb{R}$, diffuse color $\mathbf{c}^{d} \in \mathbb{R}^{3}$, and specular features $\mathbf{f} \in \mathbb{R}^{4}$:
\begin{equation}
    \tau = \exp(\tilde{\tau}),\quad \mathbf{c}^{d}=\sigma(\tilde{\mathbf{c}}^{d}),\quad\mathbf{f}=\sigma(\tilde{\mathbf{f}}).
\end{equation}

After alpha composition as in Eq.~\eqref{alpha}, the diffuse color and specular features are concatenated with the direction and fed into a deferred MLP $\mathcal{G}$ to obtain the final color:
\begin{equation}\label{defer}
    \hat{\mathbf{C}}(\mathbf{r}) = \sum\nolimits_{i=1}^{n}\omega_{i}\mathbf{c}_{i}^{d} + \mathcal{G}(\mathbf{F},\mathbf{d}),\quad\mathbf{F}=\left[\sum\nolimits_{i = 1}^{n}\omega_{i}\mathbf{c}_{i}^{d}, \sum\nolimits_{i = 1}^{n}\omega_{i}\mathbf{f}_{i}\right].
\end{equation}

After the training, MERF performs a full-resolution rendering on all images to determine the areas where the opacity and weights exceed a predefined threshold. These areas are then stored as the occupied space, preserving corresponding volumetric density, diffuse colors, and specular colors. The information is utilized for subsequent real-time rendering on the web.

\section{Method}
Based on the model combining sparse grids and triplanes, we propose a method for high-quality, real-time rendering suitable for oblique photography. Section~\ref{section:4.1} introduces a novel explicit two-dimensional occupancy plane and its optimization to obtain a compact and detailed representation for efficient sampling. Section~\ref{section:4.2} proposes a novel regularization term to address the view extrapolation issue for oblique photography with limited tilt angles. The entire optimization process is presented in Section~\ref{section:4.3}. Section~\ref{section:4.4} explains the rapid baking of scene features for real-time rendering from arbitrary viewpoints based on the proposed model.

\subsection{Occupancy Plane}
\label{section:4.1}
For reconstruction from aerially captured data, it is beneficial to align the $xy$-plane of the world coordinate system with the ground, so that the $z$-axis is perpendicular to the ground and points upward. With this configuration, we make two primary assumptions:
\begin{enumerate}[leftmargin=*]
    \item For any point on the $xy$-plane, elements in the scene are contained within a single continuous interval along the $z$-axis, i.e., no object exists isolated in mid-air;
    \item As the $z$-value increases, the occupancy within the scene gradually becomes sparser, signifying a decrease in elements or structures at higher elevations.
\end{enumerate}

\begin{wrapfigure}{r}[1em]{0.5\columnwidth}
\vspace*{-1.2em}
\hspace*{-2em}
\includegraphics[width=0.5\columnwidth]{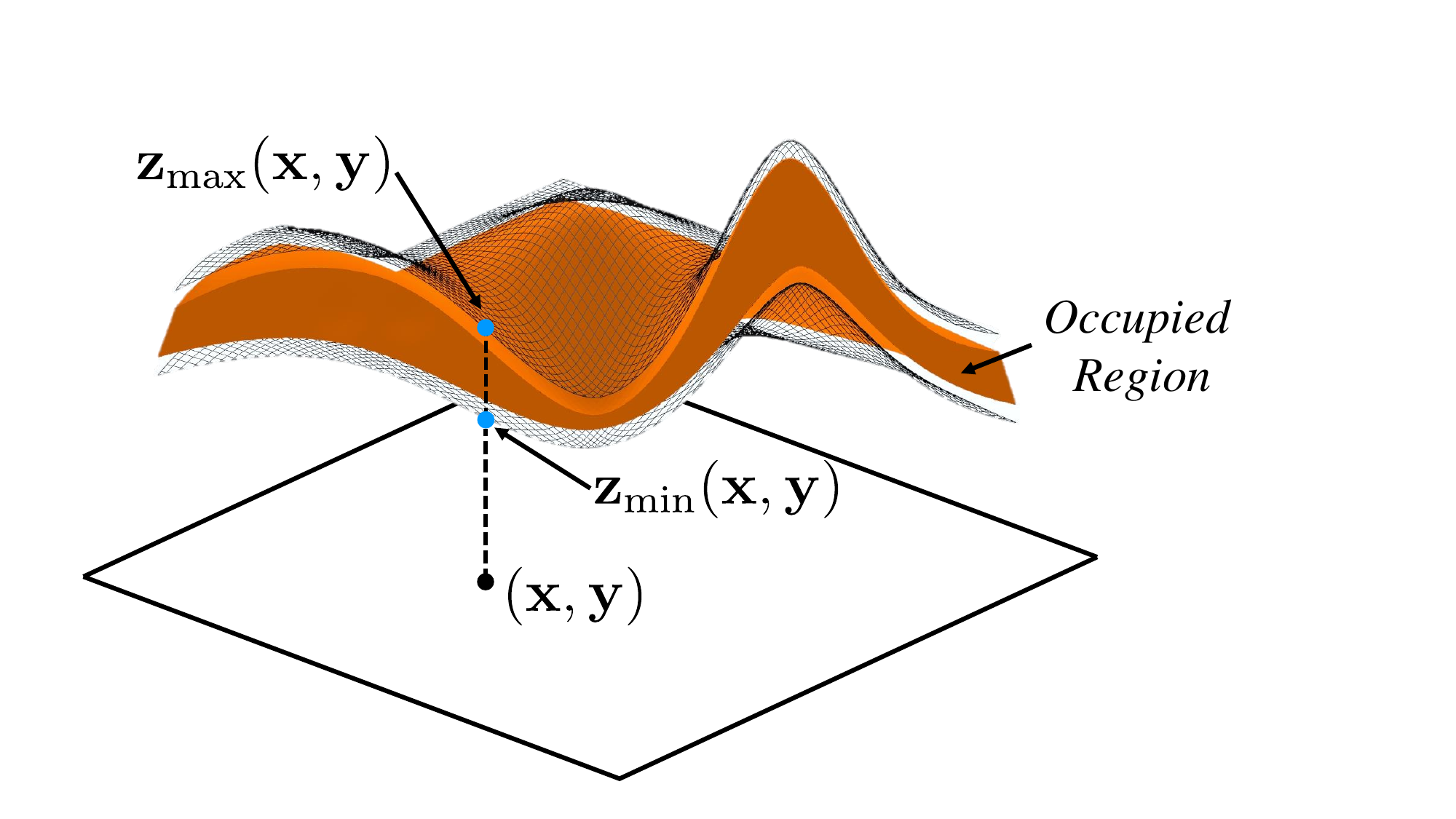}
\vspace*{-0.8em}
\end{wrapfigure}
Based on these assumptions, we represent the occupied space as a sandwiched region between two height field surfaces over the ground, corresponding to the lower and upper bounds of the $z$ coordinates, respectively (see the inset figure). 
To parameterize the occupied space, we sample the $xy$-plane with a high-resolution $M$$\times$$M$ grid, and store the heights  $\zmin(x,y)$ and $\zmax(x,y)$ of the two surfaces at each sample grid point $(x,y)$, resulting in a representation $\mathcal{P}_{o} \in \mathbb{R}^{M\times M\times 2}$. We refer to this as the occupancy plane in the following. The heights $\zmin(x,y)$ and $\zmax(x,y)$ define a continuous occupied internal $\mathcal{I}_{(x,y)} = [\zmin(x,y), \zmax(x,y)]$ for the $z$ coordinates corresponding to each grid point $(x,y)$. We set the probability of occupancy outside this interval to be zero, so that we can exclude the points outside this interval when computing the color for a ray. To maximize memory efficiency, we would like to compress the interval as much as possible. Thus, we introduce a loss function term to penalize the span of the interval:
\begin{equation}
    \mathcal{L}_{\textrm{occ}} = \sum\nolimits_{(x,y) \in \mathcal{S}}(\zmax(x,y)- \zmin(x,y))^{2},
    \label{eq:OccLoss}
\end{equation} 
where $\mathcal{S}$ denotes the set of grid points for the occupancy plane.

However, simply compressing the occupancy intervals using $\mathcal{L}_{\textrm{occ}}$ can affect the reconstruction quality if significant elements in the scene are left outside the sandwiched region. To avoid this issue, we integrate the occupancy plane into the volume rendering process, to ensure the final occupancy intervals are sufficient to represent the scene. Specifically, for the vertical line over a sample grid point $(x,y)$ in the occupancy plane, we derive a differentiable occupancy function for the points along the line based on their $z$ coordinates and the occupancy interval $[\zmin(x,y), \zmax(x,y)]$ (below we ignore the arguments $(x,y)$ for $\zmin$ and $\zmax$ to simplify the presentation):
\begin{equation}
M_{(x,y)}(z;\mathcal{P}_{o}) =
\begin{cases}
    \quad \ 1 & \text{if } z \in [z_{\min} + \epsilon, z_{\max} - \epsilon], \\
    \quad \ 0 & \text{if } z \in (-\infty, z_{\min}) \cup (z_{\max}, \infty), \\
    {(z - z_{\min})^q}/{\epsilon^q} & \text{if } z \in [z_{\min}, z_{\min} + \epsilon], \\
    {(z_{\max} - z)^q}/{\epsilon^q} & \text{if } z \in [z_{\max} - \epsilon, z_{\max}].
\end{cases}
\label{eq:OccupancyFunc}
\end{equation}
Here we use a threshold $\epsilon$ to introduce two buffer zones $[z_{\min}, z_{\min} + \epsilon]$ and $[z_{\max} - \epsilon, z_{\max}]$ near the endpoints of $[\zmin, \zmax]$, where the occupancy function transitions smoothly and monotonically from 0 to 1 using power functions. 
We set the exponent parameter $q = 2$ in this paper. During volume rendering, for a sample point $\mathbf{p}_i=(x_i,y_i,z_i)$ on a ray $\mathbf{r}$, we project it onto the XY plane and find the nearest sample grid point $(\overline{x}_i, \overline{y}_i)$ to the projection $(x_i, y_i)$, and obtain a value for $\mathbf{p}_i$ using the occupancy function at $(\overline{x}_i, \overline{y}_i)$:
\begin{equation}
    M_{(x_i,y_i)}(z_i;\mathcal{P}_{o}) = M_{(\overline{x}_i,\overline{y}_i)}(z_i;\mathcal{P}_{o}).
    \label{eq:OccValue}
\end{equation}
The value is then combined with the weight from volume rendering, to determine the contribution of the feature at $\mathbf{p}_i$ to the final color
\begin{equation}
    \hat{\mathbf{C}}(\mathbf{r}) = \sum\nolimits_{i=1}^{n} M_{(x_i,y_i)}(z_i;\mathcal{P}_{o})\omega_{i}\mathbf{c}_{i}.
\end{equation}
This is used to define a photometric loss $\mathcal{L}_{\textrm{rgb}}$ that penalizes the deviation between the predicted color and the ground truth (see Eq.~\eqref{eq:rgbloss}). 
The occupancy value in~\eqref{eq:OccValue} correlates the predicted color with the occupancy interval, such that the photometric loss suppresses further compression of the occupancy interval when its endpoints with high weights, helping to achieve both rendering quality and memory efficiency.
In addition, the occupancy value improves the efficiency of computation, as it can be utilized to filter out sampling points outside the occupied region.
This significantly reduces the number of sampling points that need to be evaluated. 

\subsection{Smoothness Regularization for View-dependent Color}
\label{section:4.2}

\begin{figure}[t]
    \centering
    \begin{subfigure}[b]{0.23\textwidth}
        \centering
        \includegraphics[width=\textwidth]{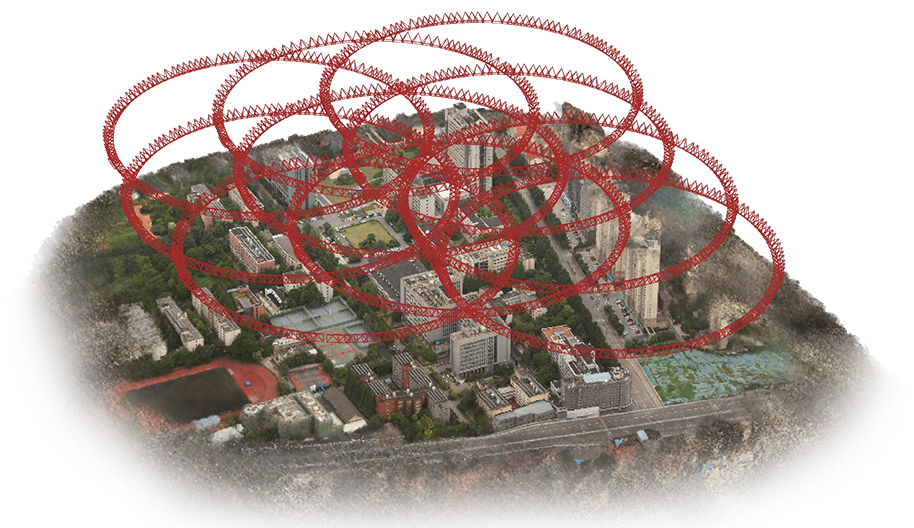}
        \caption{surround-style photography}
        \label{fig:dataset_1}
    \end{subfigure}
    \hfill 
    \begin{subfigure}[b]{0.23\textwidth} 
        \centering
        \includegraphics[width=\textwidth]{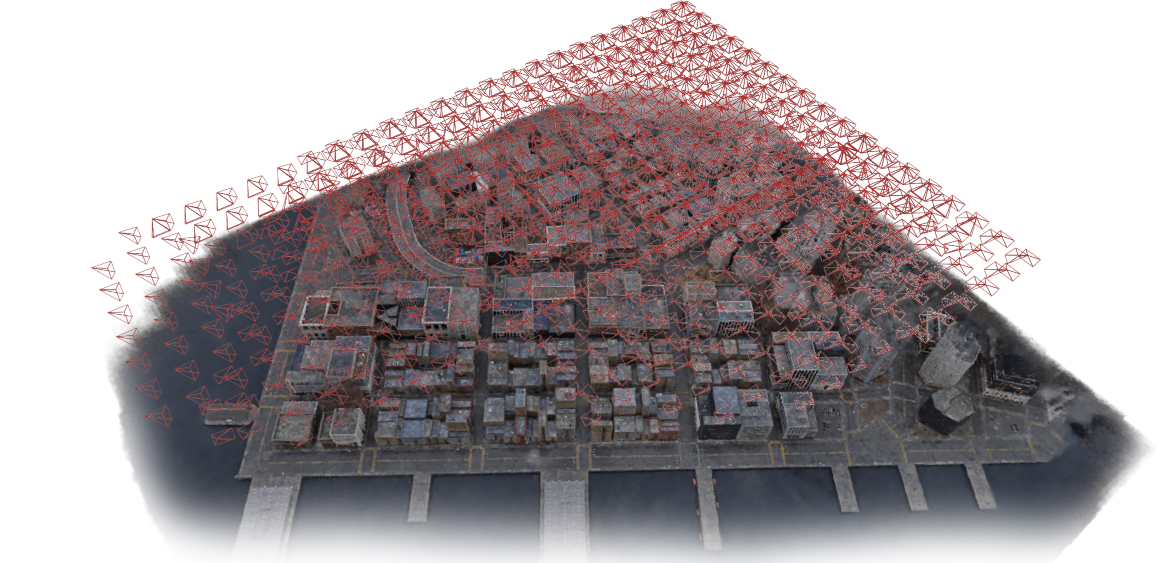}
        \caption{grid-style photography}
        \label{fig:dataset_2}
    \end{subfigure}
    \caption{Camera trajectories from two oblique photography methods.}
    \label{fig:dataset}
\end{figure}

\begin{figure}[t]
    \centering
    \begin{subfigure}[b]{0.1\textwidth}
        \centering
        \includegraphics[height=3cm]
        {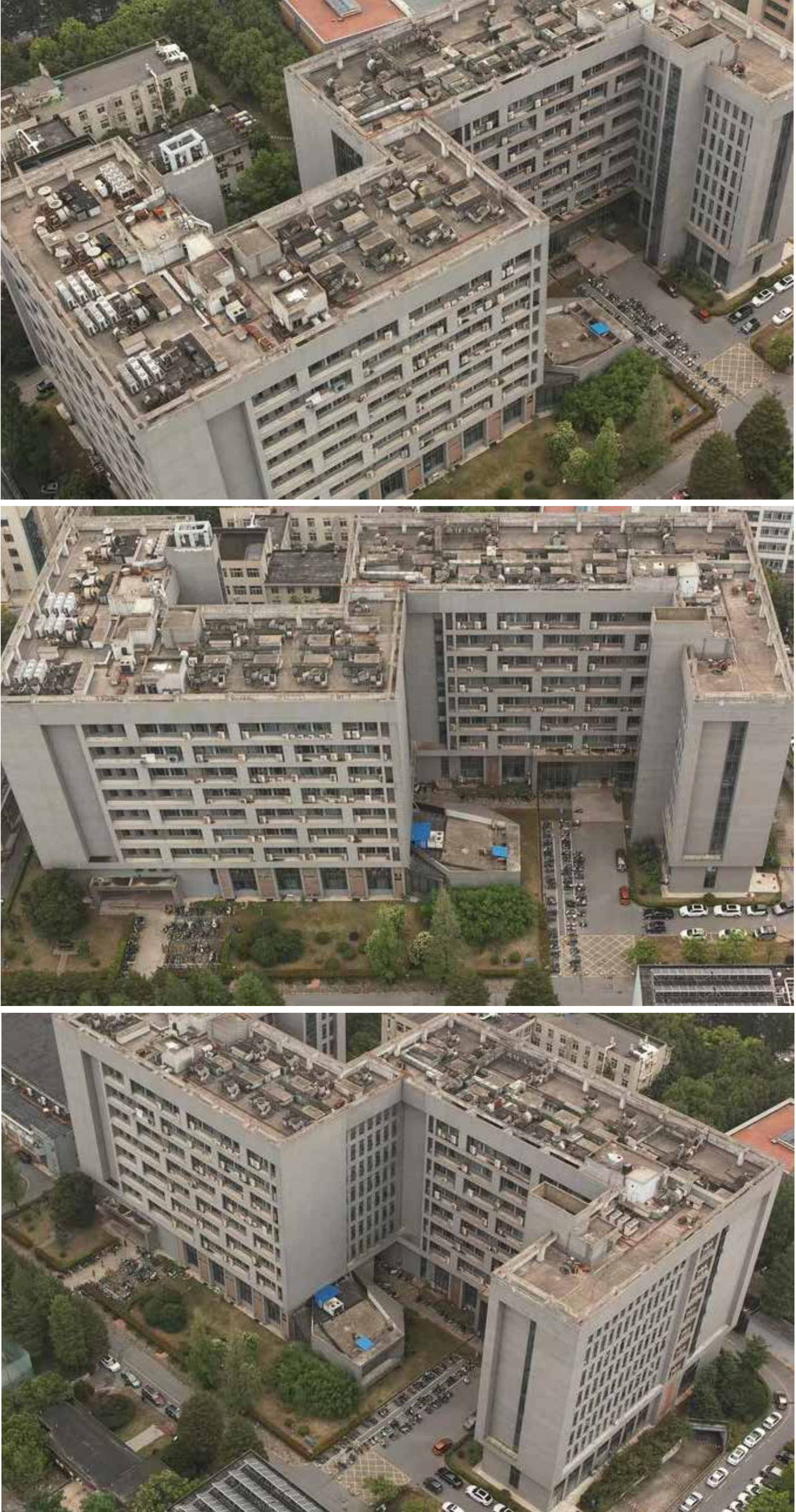} 
        \caption{Train views}
        \label{fig:1}
    \end{subfigure}
    \rule{0.6pt}{3.8cm}
    \begin{subfigure}[b]{0.15\textwidth}
        \centering
        \includegraphics[height=3cm]
        {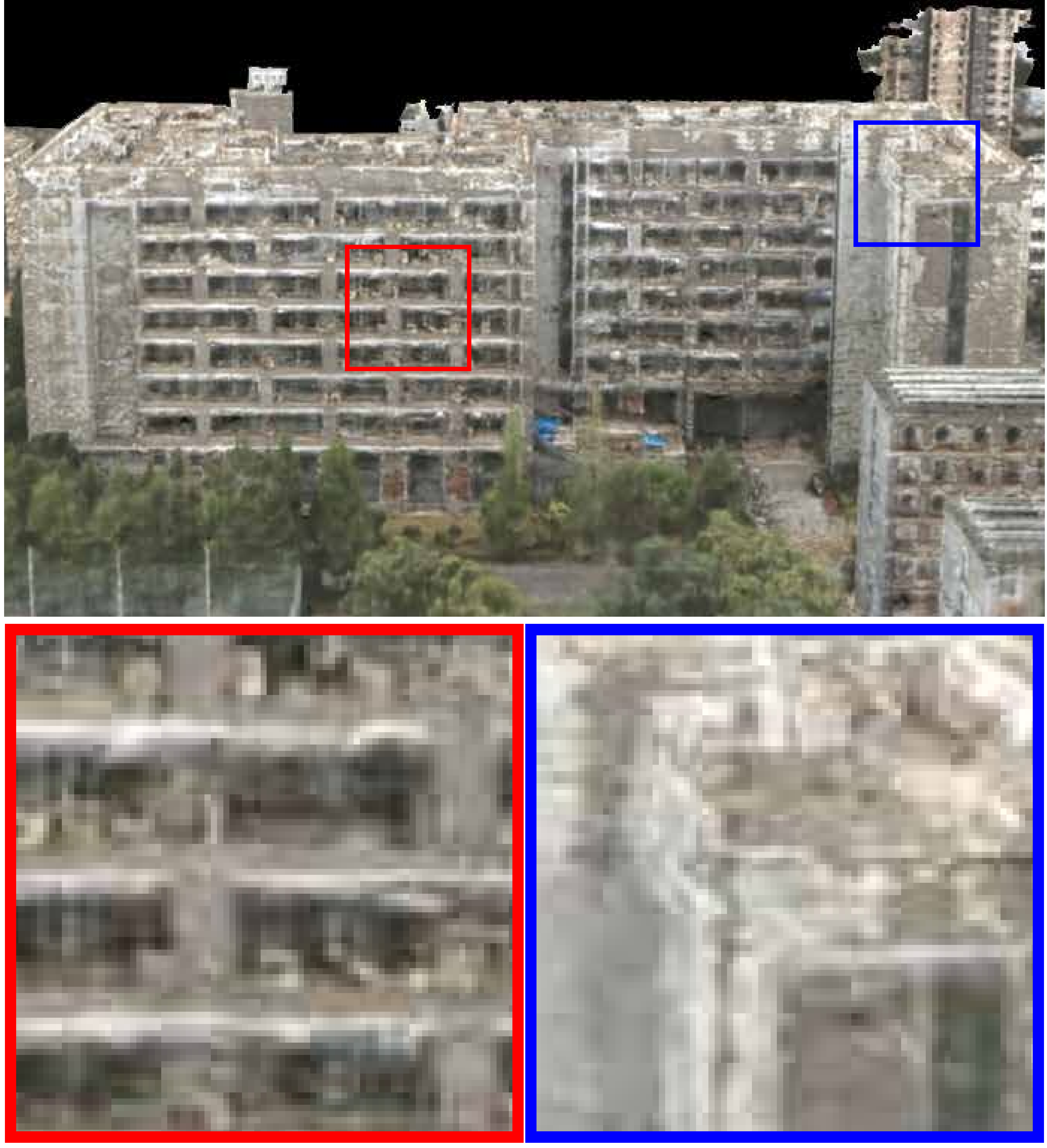} 
        \caption{no $\mathcal{L}_{\textrm{smooth}}$}
        \label{fig:2}
    \end{subfigure}
    \begin{subfigure}[b]{0.15\textwidth}
        \centering
        \includegraphics[height=3cm]
        {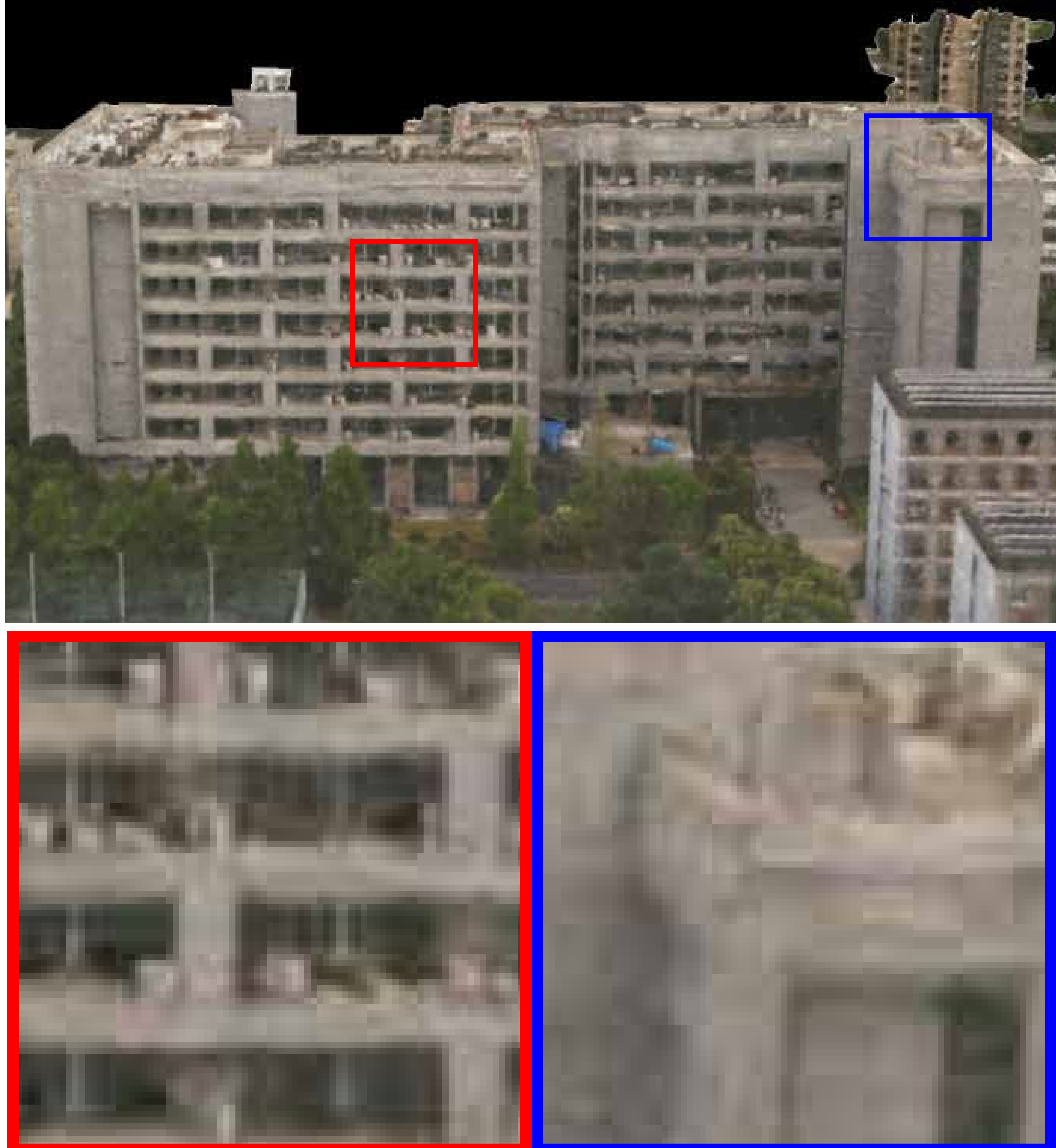}
        \caption{with $\mathcal{L}_{\textrm{smooth}}$}
        \label{fig:3}
    \end{subfigure}
    \caption{In (a), we present training views captured around a building at limited tilt angles. (b) and (c) illustrate the real-time rendering results from a novel extrapolation viewpoint without and with $\mathcal{L}_{\textrm{smooth}}$, respectively. The introduction of the smoothness regularization yields renderings that are smoother and more consistent.}
    \label{fig:three_columns}
\end{figure}

Another challenge in large-scale scene reconstruction is the extrapolation issue of view-dependent colors. As shown in Fig.~\ref{fig:dataset}, when a drone flies through the entire scene, the conventional approach involves capturing images with limited pitch angles, either by orbital capture at a consistent angle relative to the ground plane or by employing a multi-camera system to photograph the scene in a grid pattern. The limited range of pitch angles in these photography methods often results in abnormal specular colors in the reconstructed scene, particularly when the scene is observed from a horizontal or upward perspective that is outside the range of view angles in the captured images (see Fig.~\ref{fig:2}). 

In our context, the deferred MLP $\mathcal{G}$ that generates the color (see Eq.~\eqref{defer}) is supervised from a sparse set of input viewpoints. When rendering under extrapolated viewpoints, it tends to shift towards high-frequency components, leading to extreme highlights in the images. Our key observation is that adjacent viewpoints in real-world scenes often exhibit similar specular reflection colors, aligning with the local consistency seen in BRDF on smooth surfaces. We incorporate this prior into our model to encourage smoothness in view-dependent colors in unseen viewpoints.
A typical approach is to impose constraints on the network parameters to achieve this continuity~\cite{lipmlp,jacob,hoffman2019robust,moosavidezfooli2018robustness}. For example, LipschitzMLP~\cite{lipmlp} achieves smoothness of the output by constraining the Lipschitz constant of the network. However, it enforces smoothness with respect to all inputs of the MLP (i.e., the viewing direction, the diffuse color, and the specular feature), whereas we only require smoothness with respect to the viewing direction; this could lead to over-constraints of the MLP parameters and may hinder the optimization process. To address this issue, we propose a regularization that enforces the smoothness of network output concerning the viewing direction only. Specifically, for a known viewing direction $\mathbf{d} \in \mathcal{D}$ in the training set, we apply a small Gaussian perturbation to define a sampling space for the viewing direction.
\begin{equation}\label{perturbation}
    \mathcal{S}_{d} = \{\mathbf{d} + \boldsymbol{\delta}\ |\ \mathbf{d} \in \mathcal{D},\boldsymbol{\delta} \thicksim \mathcal{N}(0,\Sigma)\},
\end{equation}
where $\Sigma$ is a hyperparameter that determines the sampling range. Then, we introduce a loss to penalize large changes between the deferred MLP's outputs for $\mathbf{d}$ and the perturbed directions 
\begin{equation}
    \mathcal{L}_{\textrm{smooth}} = \sum\nolimits_{\mathbf{d} \in \mathcal{D}} \sum\nolimits_{\mathbf{s} \in \mathcal{S}_{d}} S(\mathbf{d},\mathbf{s})\|\mathcal{G}(\mathbf{F},\mathbf{s})-\mathcal{G}(\mathbf{F},\mathbf{d})\|_{2}^{2},
     \label{eq:SmoothLoss}
\end{equation}
where $S(\cdot, \cdot)$ denotes cosine similarity.
In our experiments, this regularization term significantly enhances the robustness of the deferred MLP across interpolated and extrapolated viewpoints. It effectively mitigates the instability of view-dependent colors under new viewpoints while ensuring rendering quality and 3D consistency.

\subsection{Optimization}
\label{section:4.3}

Our model is trained using a loss function written as a weighted sum:
\begin{equation}\label{eq:totalloss}
\begin{aligned}
   \mathcal{L} = & \ \lambda_1 \mathcal{L}_{\text{rgb}} +  \lambda_2 \mathcal{L}_{\text{S3IM}} + \lambda_3 \mathcal{L}_{\text{distortion}} + \lambda_4 \mathcal{L}_{\text{interval}} \\
   & + \lambda_5 \mathcal{L}_{\text{sparsity}} + \lambda_6 \mathcal{L}_{\text{entropy}} + \lambda_7 \mathcal{L}_{\text{occ}} + \lambda_8 \mathcal{L}_{\text{smooth}}.
\end{aligned}
\end{equation}
Here $\mathcal{L}_{\text{occ}}$ and $\mathcal{L}_{\text{smooth}}$ are defined in Eq.~\eqref{eq:OccLoss} and Eq.~\eqref{eq:SmoothLoss}, respectively. $\mathcal{L}_{\text{rgb}}$ is a photometric loss that penalizes the disparity between the rendered images and the ground truth images, using the Charbonnier loss~\cite{Charbonnier} as a robust norm:
\begin{equation}\label{eq:rgbloss}
   \mathcal{L}_{\textrm{rgb}} = \sum\nolimits_{\mathbf{r}\in \mathcal{R}}\sqrt{\|\mathbf{C}(\mathbf{r})-\hat{\mathbf{C}}(\mathbf{r})\|_{2}^{2}+\epsilon_{c}},
\end{equation}
where $\mathcal{R}$ is the set of training rays, and $\epsilon_{c} = 10^{-6}$ is a parameter to ensure smoothness.
$\mathcal{L}_{\text{S3IM}}$ is the S3IM loss from~\cite{xie2023s3im} to enforce structural similarity between the rendered  and input images. 
The interval loss $\mathcal{L}_{\text{interval}}$ and the distortion loss $\mathcal{L}_{\text{distortion}}$ are both adopted from~\cite{barron2022mipnerf360}; the former aligns the weight distribution predictions of the proposal MLP and the NeRF MLP to rationalize the sample point distribution, and the latter reduces floater artifacts.
$\mathcal{L}_{\textrm{sparsity}}$ is a sparsity loss defined using randomly sampled points in the occupied space to encourage lower opacity:
\begin{equation}
    \mathcal{L}_{\textrm{sparsity}} = \frac{1}{|\mathcal{P}|}\sum\nolimits_{\mathbf{p_{i}}\in \mathcal{P}} \alpha_{i}.
\end{equation}
This helps to address the issue of near-camera foggy artifacts resulting from inaccurate volume density estimation that often occurs in large scenes.
Additionally, following~\cite{kim2022infonerf}, we randomly sample a number of rays $\mathcal{R}$ from high altitude to the ground, calculate the opacity $\alpha_{i}$ of sample points $\mathbf{p_{i}}$ along each ray $\mathbf{r}$, and combine it with the previous occupancy value to derive an entropy loss for the discrete density variable in the occupied space:
\begin{equation}
    \mathcal{L}_{\textrm{entropy}} = -\frac{1}{|\mathcal{R}|}\sum\nolimits_{\mathbf{r}\in \mathcal{R}} \sum\nolimits_{\mathbf{p}_i \in \mathbf{r}} p(\mathbf{p_{i}})\log(p(\mathbf{p_{i}})),
\end{equation}
where $p(\mathbf{p_{i}}) = {M_{i}\alpha_{i}}/({\sum_{\mathbf{p}_i \in \mathbf{r}} M_{i}\alpha_{i}})$.
This encourages the occupancy probability of spatial points to approach either 0 or 1, which ensures consistency between training and real-time rendering.

During training, we initialize the weight for $\mathcal{L}_{\text{occ}}$ to a small value and gradually increase it. In this way, the training first focuses on reconstruction of the scene, and then incrementally compresses the occupied space while maintaining the reconstruction quality. Further details can be found in the supplementary materials.

\subsection{Real-time Rendering}\label{section:4.4}
After training, we store the features identified as occupied by the occupancy plane for subsequent rendering. Existing approaches~\cite{hedman2021snerg,Reiser2023SIGGRAPHMERF} construct a 3D voxel grid using the NeRF model for volume density evaluation or perform a full-resolution rendering of the training set for weight evaluation. This often requires dozens of hours for large-scale, high-resolution scenes. In contrast, we efficiently determine whether a voxel grid should store features based on the high-resolution occupancy plane obtained during training, which is completed in just a few minutes. Similar to SNeRG~\cite{hedman2021snerg} and MERF~\cite{Reiser2023SIGGRAPHMERF}, we store voxel grids as sparse blocks and the deferred MLP as floating-point arrays. For the 2D triplane, we employ a texture map with high resolution in the $xy$ direction and low resolution in the $z$ direction, according to the upper and lower bounds of the occupancy plane. This approach saves a significant amount of video memory during real-time rendering. To skip empty space efficiently, we employ 2D max pooling to obtain multiple low-resolution occupancy planes. All these features are encoded in PNG format.

After the baking process, we follow MERF~\cite{Reiser2023SIGGRAPHMERF} for real-time rendering but diverge in ray marching. Unlike the multi-resolution binary grid used to determine the current point's occupancy status, the occupancy plane directly stores the start and end positions of the point's sampling interval in the $z$ direction. As the ray traverses the blank area, it can efficiently reach the next grid point or the starting sampling position of the current grid point through bounding box intersection detection. Experimental results indicate that this sampling strategy is notably faster than previous approaches, especially when overlooking the entire scene.

\section{Experiments}
We evaluate our method in terms of rendering quality, memory consumption, and real-time rendering performance. In Section~\ref{section:5.1}, we compare our method with a series of offline novel view synthesis methods and some real-time rendering methods. In Section~\ref{section:5.2}, we validate the effectiveness of our smoothness regularization. In Section~\ref{section:5.3}, we compare the rendering quality, memory, and occupancy ratio under different resolutions of sparse feature grids.

\begin{figure*}[h]
  \centering
  \includegraphics[width=\linewidth]{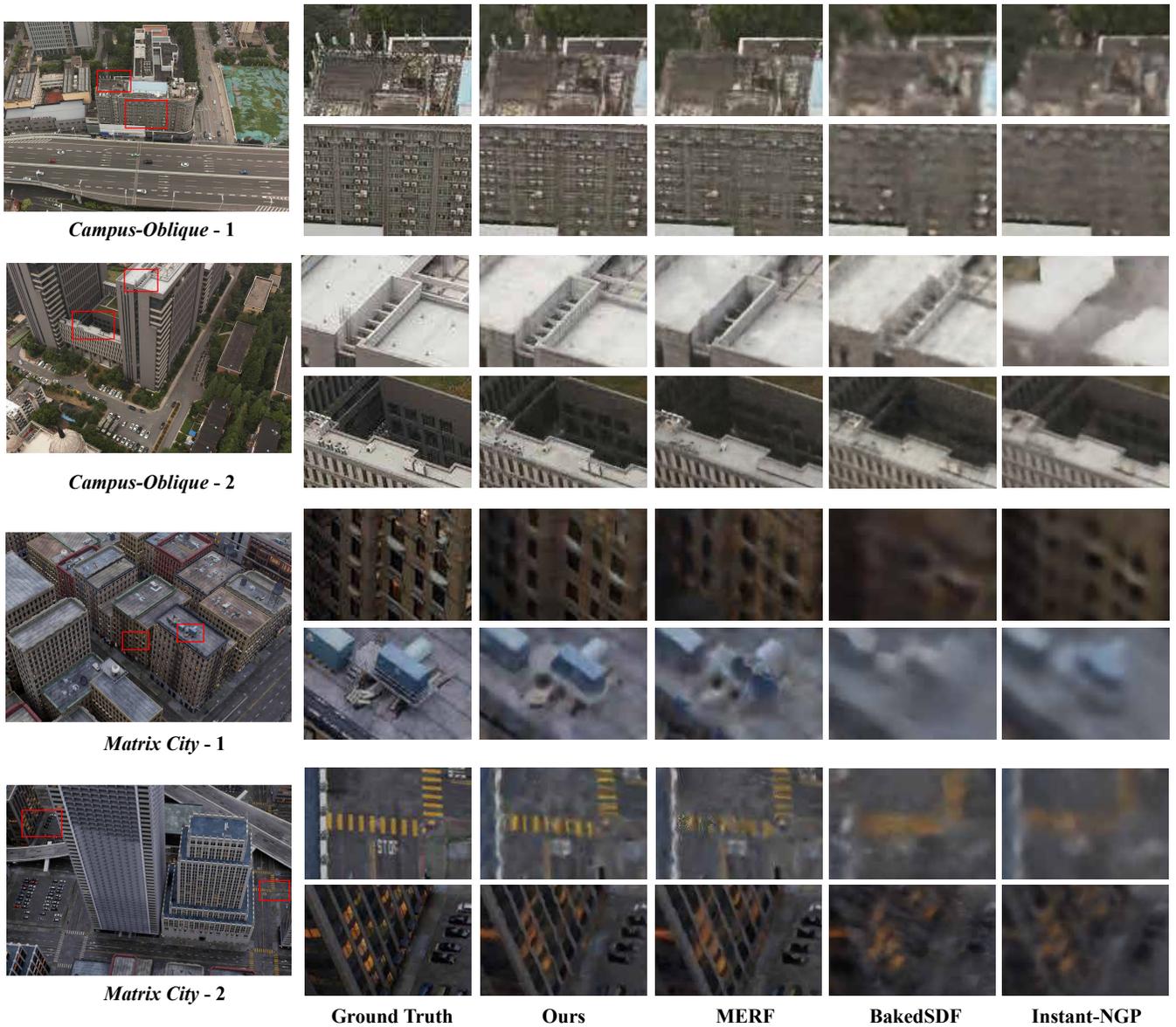}
  \caption{Comparison on rendering quality for novel view between Oblique-MERF and other methods on the \textit{Campus-Oblique} and \textit{Matrix City}~\cite{li2023matrixcity} dataset.}
  \label{fig:vis_comp}
\end{figure*}

\subsection{Real-time rendering on oblique photography dataset}\label{section:5.1}

\paragraph{Dataset and Experiment Settings. } We employ two distinct datasets for evaluation: the \textit{Matrix City}~\cite{li2023matrixcity} and \textit{Campus-Oblique} datasets. \textit{Matrix City} is a synthetic dataset, captured in a grid-style format typical of classic oblique photography. \textit{Campus-Oblique} is a real-world dataset captured by ourselves using a surround-style approach. It encompasses three distinct scenes on a university campus. Two of them cover an area of about 120,000 square meters and comprise over 1,000 images each. The third one is even more extensive, covering approximately 300,000 square meters and containing over 3,000 images. 
For each scene, we use 99\% of the images for training, and the remaining ones for testing. Our training code and baseline MERF~\cite{Reiser2023SIGGRAPHMERF} is built upon the nerfstudio framework~\cite{nerfstudio}, augmented with the tiny-cuda-nn~\cite{tiny-cuda-nn} extension. Our real-time viewer is implemented as a JavaScript web application, utilizing GLSL for rendering. We conduct comprehensive comparisons with several offline and real-time methods to evaluate the performance of our method. For offline models, we compare with established methods like NeRFacto~\cite{nerfstudio} and Instant-NGP~\cite{mueller2022instant}. For real-time models, we compare with MobileNeRF~\cite{chen2022mobilenerf} and BakedSDF~\cite{yariv2023bakedsdf}. We use MERF as the baseline for the proposed method and set the triplane resolution $R$
to 2048, the sparse grid resolution $L$ to 512 and the occupancy plane resolution $M$ to 512. 
We evaluate the rendering quality using a set of established metrics: peak signal-to-noise ratio {PSNR}, {SSIM}~\cite{Wang2004}, and {LPIPS}~\cite{zhang2018perceptual}. Additionally, we use GPU memory usage ({VRAM}), frames per second ({FPS}), and on-disk storage ({DISK}) as metrics for the efficiency of real-time rendering methods.

\begin{table}[t]
\caption{Quantitative results on \textit{Matrix City} and \textit{Campus-Oblique} datasets.}
\setlength{\tabcolsep}{3pt} 
\renewcommand{\arraystretch}{0.9} 
  \centering
    \begin{tabular}{l|ccc|ccc}
    \toprule
          & \multicolumn{3}{c|}{\textit{Campus-Oblique}} & \multicolumn{3}{c}{\textit{Matrix City}} \\
          & PSNR↑ & SSIM↑ & LPIPS↓ & PSNR↑ & SSIM↑ & LPIPS↓ \\
    \midrule
    InstantNGP & 22.49 & 0.583 & 0.540  & 23.09 & 0.612 & 0.707 \\
    Nerfacto & 22.07 & 0.599 & 0.361 & 23.40  & 0.674 & 0.443 \\
    MobileNeRF & 20.49 & 0.409 & 0.529 & 21.48 & 0.526 & 0.544 \\ 
    BakedSDF & 21.86 & 0.537 & 0.498 & 22.09 & 0.582 & 0.652 \\
    MERF & 23.44 & 0.667 & 0.299 & 24.50  & 0.679 & 0.454 \\
    Ours & \textbf{24.14} & \textbf{0.694} & \textbf{0.270} & \textbf{25.18} & \textbf{0.714} & \textbf{0.406} \\
    \bottomrule
    \end{tabular}%
  \label{tab:Comparation_quality}%
\end{table}%

\paragraph{Results.}
In Tab.~\ref{tab:Comparation_quality}, we conduct a quantitative comparison of rendering quality between our method and both offline and real-time rendering methods. Our approach not only matches offline methods in all metrics but also outperforms competing real-time rendering solutions. For real-world scenes, our method demonstrates superior rendering quality, thanks to the color network's robustness enhanced by the proposed smoothness regularization term. As shown in Fig.~\ref{fig:vis_comp}, unlike other models that blur natural scene details, our method preserves clear and high-frequency details. Our sampling space regularization, which specifically addresses scene geometry orthogonal to the ground, allows for the depiction of crisp surface details, resulting in sharper geometry compared to the baseline.

We evaluate the real-time rendering performance of our method and compare it with MobileNeRF~\cite{chen2022mobilenerf}, BakedSDF~\cite{yariv2023bakedsdf}, and MERF~\cite{Reiser2023SIGGRAPHMERF} in Tab.~\ref{tab:real-rendering}. The evaluation is carried out at 1920$\times$1080 resolution on an NVIDIA RTX 1650. Despite the inherently lower frame rates of volumetric rendering against mesh rasterization, our method excels in DISK and VRAM efficiency, while delivering exceptional real-time rendering quality as shown in Tab.~\ref{tab:Comparation_quality}. Our compact occupancy space design yields requires less storage and improves frame rates. Remarkably, our sampling approach maintains consistent rendering speeds across scales, in contrast to MERF's performance drop in large-scale scenes.

\begin{table}[t]
\caption{The performance for our model and other real-time methods.}
\setlength{\tabcolsep}{3pt} 
\renewcommand{\arraystretch}{0.85} 
  \centering
    \begin{tabular}{l|cccc}
    \toprule
          & \multirow{2}{*}{VRAM↓}  & \multirow{2}{*}{DISK↓}  & \multicolumn{2}{c}{FPS↑} \bigstrut\\
          \cline{4-5}
          & (MB)  & (MB)  & high-altitude & low-altitude \bigstrut\\
    \midrule
    MobileNeRF & 1117.0  & 399.1 & \textbf{63}  & 43 \\
    BakedSDF & 595.0 & 509.0 & 58 & \textbf{61} \\
    MERF & 185.9 & 97.7 & 8  & 31 \\
    Ours  & \textbf{108.7} & \textbf{75.1} & 32  & 42 \\
    \bottomrule
    \end{tabular}%
  \label{tab:real-rendering}%
\end{table}%

\begin{table}[t]
\caption{Effectiveness of the Smoothness Regularization.}
\setlength{\tabcolsep}{3pt} 
\renewcommand{\arraystretch}{0.9} 
  \centering
  \begin{tabular}{l|ccc|ccc}
    \toprule
    & \multicolumn{3}{c|}{Training views} & \multicolumn{3}{c}{Test views} \\
    & PSNR↑ & SSIM↑ & LPIPS↓ & PSNR↑ & SSIM↑ & LPIPS↓ \\
    \midrule
    MERF & 25.93 & 0.770 & 0.189 & 22.14 & 0.710 & 0.213 \\
    +LipMLP & 25.86 & 0.768 & 0.188 & 22.42 & 0.729 & 0.201 \\
    +Gradient & 25.97 & 0.772 & 0.187 & \textbf{22.66} & 0.732 & 0.199 \\
    \midrule
    Ours & \textbf{26.00} & \textbf{0.776} & \textbf{0.183} & \textbf{22.66} & \textbf{0.744} & \textbf{0.196} \\
    \bottomrule
  \end{tabular}
  \label{tab:extr-color}
\end{table}

\subsection{Color for extrapolation novel viewpoints}\label{section:5.2}

\begin{figure*}[h]
  \centering
  \includegraphics[width=\linewidth]{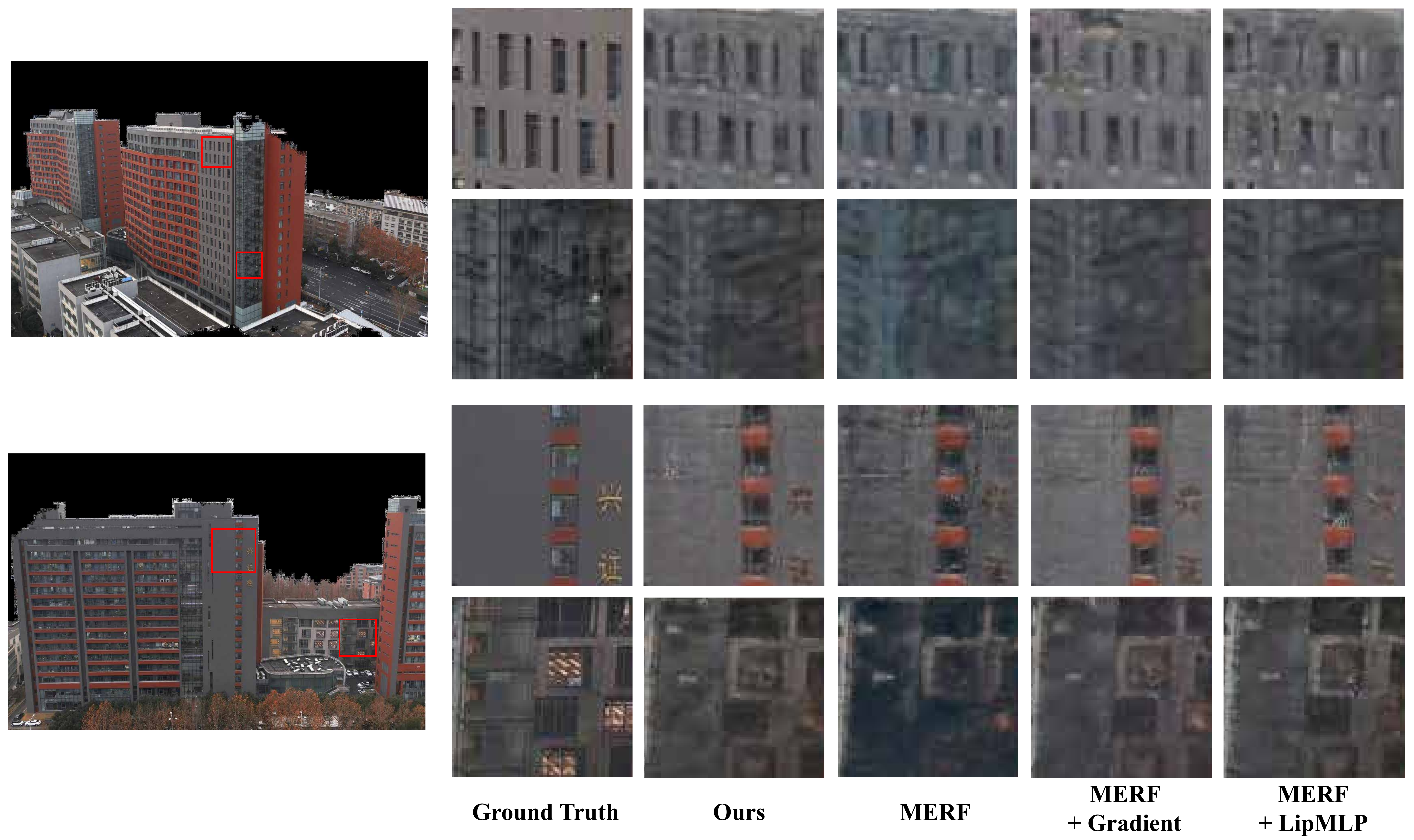}
  \caption{Rendering quality comparison between Oblique-MERF and other MERF variants for test views on the \textit{Campus-extra} dataset.}
  \label{fig:com_specular}
\end{figure*}

To validate the effectiveness of our smoothness regularization term for specular color introduced in Section~\ref{section:4.2}, we employ a novel real-world dataset named \textit{Campus-extra}. For the training set, we follow the same technique as \textit{Campus-Oblique} to capture 1486 images around a building, with a fixed tilt angle to the ground at approximately 60 degrees. For the test set, we simulate extrapolated viewpoints by capturing 140 images with tilt angles at around 25 degrees. To minimize the impact of unseen scenes, we adopt the NerfBusters~\cite{Nerfbusters2023} protocol to mask test images to include only regions observed during training.

Tab.~\ref{tab:extr-color} showcases our method's rendering performance, alongside a series of MERF variants under both interpolated and extrapolated viewpoints. In MERF+LipMLP, we replace the deferred MLP with a Lipschitz MLP and corresponding regularization~\cite{lipmlp}. In MERF+Gradient, we penalize the $\ell_2$-norm of the specular color's gradient with respect to the viewing direction. Fig.~\ref{fig:com_specular} qualitatively demonstrates the results of these methods at test views. We find that merely ensuring the network's Lipschitz continuity slightly enhances the test view performance but can negatively impact the training views. This is potentially because it enforces smoothness on all network inputs including diffuse and specular features, which is redundant and hinders the optimization. Furthermore, applying regularization on the specular color's gradient markedly improves test set performance but can disrupt some views' optimization, occasionally resulting in undesirable artifacts. Compared to these approaches, our method consistently excels on both the training and test sets, achieving an approximate improvement of 0.5dB PSNR over the baseline and producing visually superior results.

\subsection{Abalation Study on Occupancy Space}\label{section:5.3}

Tab.~\ref{tab:res} compares our method with the baseline MERF in rendering quality, memory usage, and occupancy ratio ({OR}) on one scene of the \textit{Campus-Oblique} dataset. To more clearly illustrate the compactness of the occupied space optimized by our sampling strategy, we exclusively employed sparse feature grids at resolutions from 512 to 2048, omitting the use of triplane. During the training process, occupancy planes matching the sparse feature grids' resolution are utilized. After training, we extract a 3D occupancy grid (requiring roughly 2 to 10 minutes), contrasting it with the one obtained from MERF's baking process (about 2 to 7 hours). Our approach consistently yields a lower OR (approximately $60\%$ of MERF's), enhancing memory efficiency and rendering quality across all tested resolutions. This demonstrates that the regularization term introduced in Section~\ref{section:4.1} effectively eliminates redundant sampling areas within objects or in the air, aligning closely with scene geometry.

\begin{table}[t]
\setlength{\tabcolsep}{1.5pt} 
\renewcommand{\arraystretch}{0.9} 
\caption{Comparison of rendering quality and memory usage. VRAM capacity is denoted in megabytes (MB).}
  \centering
  \small
    \begin{tabular}{l|ccc|ccc|ccc}
    \toprule
    Spatial & \multicolumn{3}{c|}{512\textsuperscript{3}} & \multicolumn{3}{c|}{1024\textsuperscript{3}} & \multicolumn{3}{c}{2048\textsuperscript{3}} \\
     Res.     & PSNR↑  & VRAM↓  & OR↓    & PSNR↑  & VRAM↓  & OR↓    & PSNR↑  & VRAM↓  & OR↓ \\
    \midrule
    MERF  & 21.83 & 94.8 & 2.54\% & 23.35 & 441.0 & 1.97\% & 24.18 & 2478.0 & 1.34\% \\
    Ours  & \textbf{22.84} & \textbf{61.3} & \textbf{1.49\%} & \textbf{24.54} & \textbf{322.5} & \textbf{1.27\%} & \textbf{25.24} & \textbf{1605.0} & \textbf{0.89\%} \\
    \bottomrule
    \end{tabular}%
  \label{tab:res}%
\end{table}%

\section{Conclusion}
We introduced Oblique-MERF, a compact and robust model optimized for real-time NeRFs in large-scale scenes, specially designed for oblique photography. Our key contribution is an innovative adaptive two-dimensional occupancy plane that is integrated with volume rendering and optimized during training. This approach ensures a balance between memory efficiency and rendering quality, while avoiding prolonged baking after training. We also introduced a smoothness regularization term for specular color relative to viewing directions, producing more natural rendering results for novel extrapolated viewpoints. Compared to existing real-time rendering techniques, Oblique-MERF delivers superior rendering quality and lower memory usage, achieving higher real-time frame rates.

Although our method improves upon the baseline in terms of sampling space and real-time rendering rates, it still performs volume rendering similar to MERF~\cite{Reiser2023SIGGRAPHMERF}. Compared to methods based on mesh rasterization, it performs slightly worse in real-time frame rates and faces challenges on devices with weaker GPUs. Furthermore, scalability remains an issue to be addressed. Oblique photography often requires reconstructing larger scenes. While our proposed occupancy plane offers a more efficient representation than 3D grids, resolution limitations still exist. Adopting divide-and-conquer strategies, similar to Block-NeRF~\cite{Tancik_2022_CVPR_Block-NeRF} and Mega-NeRF~\cite{Turki_2022_CVPR_Mega-NERF}, could be a viable solution for enhancing representation ability of our model. Additionally, our smoothness prior mitigates artifacts in extrapolated viewpoints to some extent but does not faithfully reproduce view-dependent colors in the scene. Integrating our method with physically-based rendering to accurately simulate the changes in specular color with viewing direction represents a future research direction.

\begin{acks}
This research was supported by the National Natural Science Foundation of China (No.62122071, No.62272433), the Youth Innovation Promotion Association CAS (No. 2018495) and the Fundamental Research Funds for the Central Universities (No. WK3470000021).
\end{acks}

\bibliographystyle{ACM-Reference-Format}
\bibliography{paper-cite}

\end{document}


\title{Oblique-MERF: Revisiting and Improving MERF for Oblique Photography}
\maketitle

\appendix

\section{OPTIMIZATION AND MODEL ARCHITECTURE}
Both MERF and Oblique-MERF models follow the same training hyperparameters and architectural designs. In the experiments in sections 5.1 and 5.3, we conduct training over 80,000 iterations with a batch size of 32,768 pixels. The experiments in section 5.2 undergo training for 50,000 iterations with the same batch size. We utilize the Adam optimizer with a learning rate that exponentially decays. The occupancy plane's learning rate drops from $5\times 10^{-5}$ to $1\times 10^{-5}$, and for other variables, from $1\times 10^{-2}$ to $1\times 10^{-3}$. The Adam optimizer's hyperparameters $\beta_1$, $\beta_2$ and $\epsilon$ are set to 0.9, 0.99 and $1\times 10^{-15}$ respectively. Training losses are initially balanced with $\lambda_{1} = 1.0$, $\lambda_{2} = 1.0$, $\lambda_{3} = 0.01$, $\lambda_{4} = 1.0$, $\lambda_{5} = 0.05$, $\lambda_{6} = 0.001$, $\lambda_{7} = 0.0$, $\lambda_{8} = 0.1$ at the begining. From the 10,000th iteration onwards, $\lambda_{7}$ is adjusted to $1\times 10^{-4}$, and every 2,000 iterations thereafter, we increase this specific loss weight by a factor of 1.5 up to 0.2. The parameters $\Sigma_1$ and $\Sigma_2$ are set to 0.3 in Eq.(9). We sample 100 rays for smooth loss, $2^{10}$ rays for smooth loss, and $2^{14}$ samples for sparsity loss. 

For the decoder MLP for density and color, we utilize a 3-layer MLP with 64 hidden units per layer. This MLP generates an 8-dimensional output vector, including density, diffuse colors, and a 4-dimensional vector for view-dependent features. In developing our deferred view-dependency model, we employ a 3-layer MLP with 16 hidden units, and the viewing directions are encoded using four frequencies like in MERF~\cite{Reiser2023SIGGRAPHMERF}. Similar to Mip-NeRF 360~\cite{barron2022mipnerf360}, our method incorporates hierarchical sampling across three levels, necessitating the use of two Proposal-MLPs. Each Proposal MLP comprises two layers with 64 hidden units and utilizes hash encoding.

\section{COMPARATIVE METHOD SETTINGS}

For MobileNeRF~\cite{chen2022mobilenerf}, We initialize a $192 \times 192 \times 192$ grid to generate polygonal meshes while adhering to default parameters for other setups. We use the open-source version\footnote{https://github.com/hugoycj/torch-bakedsdf} for BakedSDF~\cite{yariv2023bakedsdf}, setting the batch size to 16,384 and conducting training in two phases: 20,000 and 50,000 epochs, respectively. We employ the official implementation of Instant NGP~\cite{mueller2022instant}, completing training over 100,000 epochs. For handling large scenes, we opt for the "big.json" configuration file included with the released code. Nerfacto is used with its default settings with a batch size of 65,536, and a training duration of 40,000 epochs.

\section{DATASETS}

To mitigate variations in lighting and shadows within the scene, our \textit{Campus-Oblique} dataset is captured under consistent, cloudy conditions. We adopt a surround-style capturing method, known for its higher overlap rate, which is superior to traditional grid-style capturing techniques. The datasets were recorded at altitudes ranging from 150 to 180 meters. Meanwhile, our \textit{Campus-extra} dataset is captured at approximately 150 meters altitude with a 60-degree tilt angle to the ground, encompassing an area of 120,000m² for the training set. For the test set, we captured two high-rise buildings, each 80 meters tall, within the scene at altitudes of 40 to 80 meters and tilt angles ranging from 10 to 20 degrees to the ground.

Camera poses were estimated using colmap~\cite{colmap}, employing a vocabulary tree for feature matching. This process was augmented with a hierarchical mapper and several rounds of triangulation and bundle adjustment to refine the camera pose estimations.
\bibliographystyle{ACM-Reference-Format}
\bibliography{paper-cite}